\documentclass[letterpaper, 10 pt, conference]{ieeeconf}  

\IEEEoverridecommandlockouts                              

\usepackage{graphicx}
\usepackage{amsmath}
\usepackage{float}
\usepackage{url}
\usepackage{soul, color, xcolor}
\usepackage{booktabs} 
\usepackage{multirow}
\usepackage{subfigure}

\title{\LARGE \bf
High-Precision Transformer-Based Visual Servoing for Humanoid Robots in Aligning Tiny Objects
}

\author{Jialong Xue, Wei Gao, Yu Wang, Chao Ji, Dongdong Zhao, Shi Yan and Shiwu Zhang
\thanks{This work was supported in part by the National Natural Science Foundation of China under Grants U22B2040 and U21A20119, in part by the Major Project of Anhui Province's Science and Technology Innovation Breakthrough Plan (202423h08050003), and in part by the Chinese Scholartree Ridge State Key Lab under the open grant 8KD005(2023)-5. (Corresponding authors: Wei Gao; Yu Wang.)}%
\thanks{Jialong Xue, Wei Gao, Yu Wang, Chao Ji and Shiwu Zhang are with the Institute of Humanoid Robots, Department of Precision Machinery and Precision Instrumentation, University of Science and Technology of China, Hefei, Anhui 230026, China. {\tt\footnotesize weigao@ustc.edu.cn; wangyuustc@ustc.edu.cn}}%
\thanks{Dongdong Zhao and Shi Yan are with the School of Information Science and Engineering, Lanzhou University, Lanzhou, Gansu 730000, China.}
}

\begin{document}

\maketitle
\thispagestyle{empty}
\pagestyle{empty}

\begin{abstract}

High-precision tiny object alignment remains a common and critical challenge for humanoid robots in real world. To address this problem, this paper proposes a vision-based framework for precisely estimating and controlling the relative position between a handheld tool and a target object for humanoid robots, e.g., a screwdriver tip and a screw head slot. By fusing images from the head and torso cameras on a robot with its head joint angles, the proposed Transformer-based visual servoing method can correct the handheld tool's positional errors effectively, especially at a close distance.
Experiments on M4-M8 screws demonstrate an average convergence error of $0.8$-$1.3$ mm and a success rate of $93\%$-$100\%$. Through comparative analysis, the results validate that this capability of high-precision tiny object alignment is enabled by the Distance Estimation Transformer architecture and the Multi-Perception-Head mechanism proposed in this paper.

\end{abstract}

\section{INTRODUCTION}

Humanoid robots see great potential in replacing human labor for automated operations, such as assembly and maintenance tasks~\cite{chang2024insert}, due to their high degrees of freedom and flexible sensor deployment. However, as floating base systems, their random initial positions often prevent them from completing tasks through fixed motion sequences like industrial robotic arms, necessitating reliance on various sensors for feedback control. Furthermore, constrained by cost and weight limitations, humanoid robots typically employ lower-precision joint encoders and less rigid mechanical structures compared to collaborative robotic arms, which makes high-precision operation tasks quite challenging.

Visual Servoing (VS) is a feedback control technique that uses visual information to enhance the precision and versatility of a robot's motions~\cite{chaumette2021visual}. 
It involves extracting visual features from scene images and then controlling the robot to manipulate objects accordingly~\cite{cong2021combination,li2017enhanced}. In the process, more than one camera
can be used to drive the errors between current and desired positions of visual features to zero~\cite{cong2021combination,wu2022survey,cong2022new,zhao2021image}. 
However, many existing techniques rely on conventional visual servoing with manually designed markers, which prevents them from being stably applied in general scenarios. 
With the advancement of machine learning technologies, vision-based imitation learning has emerged in recent years as a novel visual servoing approach, such as BeT~\cite{shafiullah2022behavior}, RT-1~\cite{brohan2022rt} and ACT~\cite{zhao2023learning}, witnessing widespread applications to humanoid robots~\cite{fu2024mobile}. Although imitation learning based control systems have shown strong scene adaptability, they perform poorly in high-precision operation tasks. According to~\cite{zhao2023learning}, ACT can only achieve a $20\%$ success rate in the Socket Peg Insertion task (5mm tolerance) and a $20\%$ success rate in the Thread Velcro Insertion task (about 3mm tolerance), while BeT and RT-1 have never succeeded in those tasks. On the other hand, imitation learning methods need to collect sufficiently large amount of data to achieve better results, the data collection process is complex, expensive and time-consuming. Therefore, more interpretable visual servoing methods for high-precision operation tasks are desired.

\begin{figure}[t!]
    \centering
    \includegraphics[width=1.0\columnwidth]{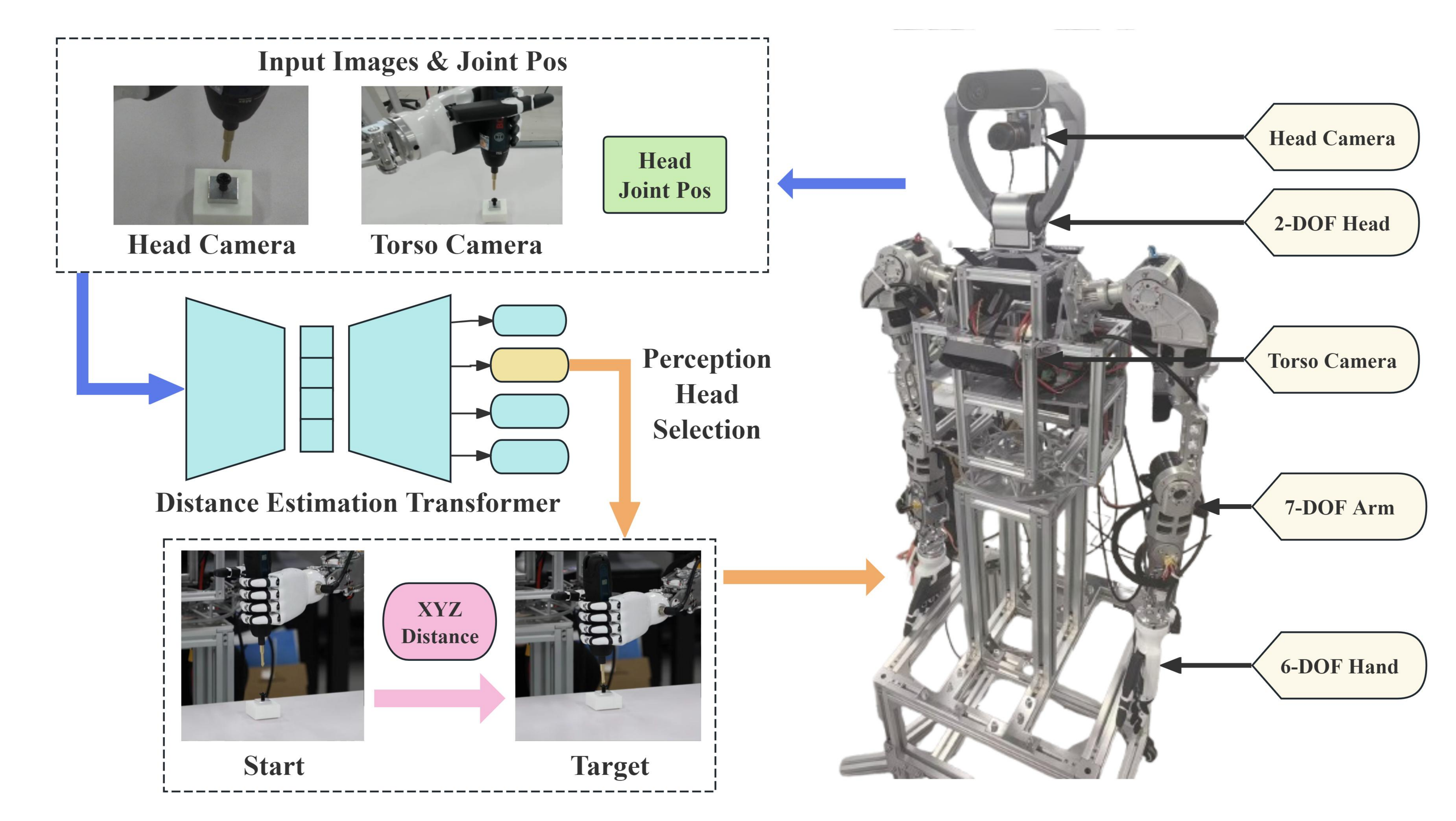}
    \caption{Overview: Tiny object alignment tasks by humanoid robots based on Multi-Perception-Head Distance Estimation Transformer.}
    \label{fig:overall}
\end{figure}

For high-precision operation, two commonly applied visual servo control schemes are Position-Based Visual Servoing (PBVS) and Image-Based Visual Servoing (IBVS). The former obtains the target object's position through images, calculates the end-effector's position through kinematics, and then computes the error in between, while the latter directly extracts the error between the target object's position and the end-effector's position from images, without relying on the kinematics. Compared to PBVS, IBVS demonstrates greater robustness to mechanical structure  and calibration errors~\cite{cong2023review}. 
Consequently, some researchers have applied IBVS to alignment tasks~\cite{chang2024insert, zheng2017peg, songautomated}, where auxiliary information provided by force sensors is also used for improved precision and success rate. 
However, high-precision force sensors are often of high cost, and even more critical is the difficulty of obtaining useful force feedback data from tiny objects. Besides, these methods rely heavily on depth information from RGB-D cameras, which again cannot perform well on locating tiny objects or features, due to the corresponding noise level of approximately $1$ cm~\cite{realsense}. 
For tiny object alignment, \cite{angelopoulos2023high} proposes a keypoint-based method with a carefully arranged camera group at close distance. However, the cameras are mounted parallel to the $x$-$z$ and $y$-$z$ planes, which is not suitable for humanoid robots in common working conditions.

Therefore, this paper focuses on how to complete tiny object alignment tasks with high precision and general scene adaptability, using only the visual sensors installed on humanoid robots. On the other hand, Transformer has been widely applied in image processing tasks~\cite{dosovitskiy2020image, carion2020end}, due to its powerful feature extraction capability and good scalability~\cite{vaswani2017attention}. It can fuse images with other data simultaneously~\cite{zhao2023learning}, particularly suitable for high-degree-of-freedom humanoid robots equipped with multiple cameras. As a result, this paper proposes a Transformer-based IBVS system to achieve high-precision visual servoing for tiny object alignment with satisfying precision and success rates, as shown in Fig.~\ref{fig:overall}.

Specifically, this paper focuses on the scheme for high-precision translational alignment, while pose alignment remains for future work as discussed in subsection~\ref{Data Collection}.
The main contributions of the paper are:
\begin{itemize}
\item A Distance Estimation Transformer architecture for fusing multiple images and joint angles in visual servoing, which helps achieve high-precision tiny object alignment on a physical humanoid robot, without relying on manually designed visual features or markers.

\item A Multi-Perception-Head mechanism that enhances network performance at close distance during visual servoing by scheduling output gains, which achieves significantly higher precision convergence.
\end{itemize}

The remainder of the paper is organized as follows: Section \ref{METHODOLOGY} presents the control framework as well as the process of data collection, Section \ref{RESULT} discusses the experimental results, and Section \ref{CONCLUSION AND LIMITATION} points out the limitations and future directions of this paper.

\section{METHODOLOGY}
\label{METHODOLOGY}

This section introduces the method for controlling the humanoid robot in Fig.~\ref{fig:overall} to perform tiny object alignment. 
The physical robot integrates dual 7-degree-of-freedom (DOF) robotic arms and a 2-DOF head. The motor at each joint incorporates a 14-bit output encoder, enabling precise arm motion control with a position accuracy of approximately $0.4$ mm. Each arm is equipped with a 6-DOF five-fingered dexterous hand for fine manipulation. The visual perception system consists of two cameras: a head-mounted HikRobot MV-CS020-10UC with a $12$ mm lens, providing a diagonal field of view (FOV) of $40^{\circ}$ and a maximum resolution of $1624 \times 1240$ pixels, and a torso-mounted Logitech C1000e camera, offering a $78^{\circ}$ diagonal FOV with $1920 \times 1080$ pixel resolution. The computation relies on an NVIDIA AGX Orin with $275$ TOPS AI performance. 

\subsection{Overall Architecture}
\label{Overall Architecture}

\begin{figure*}[!t]
    \centering
    \includegraphics[width=1.7\columnwidth]{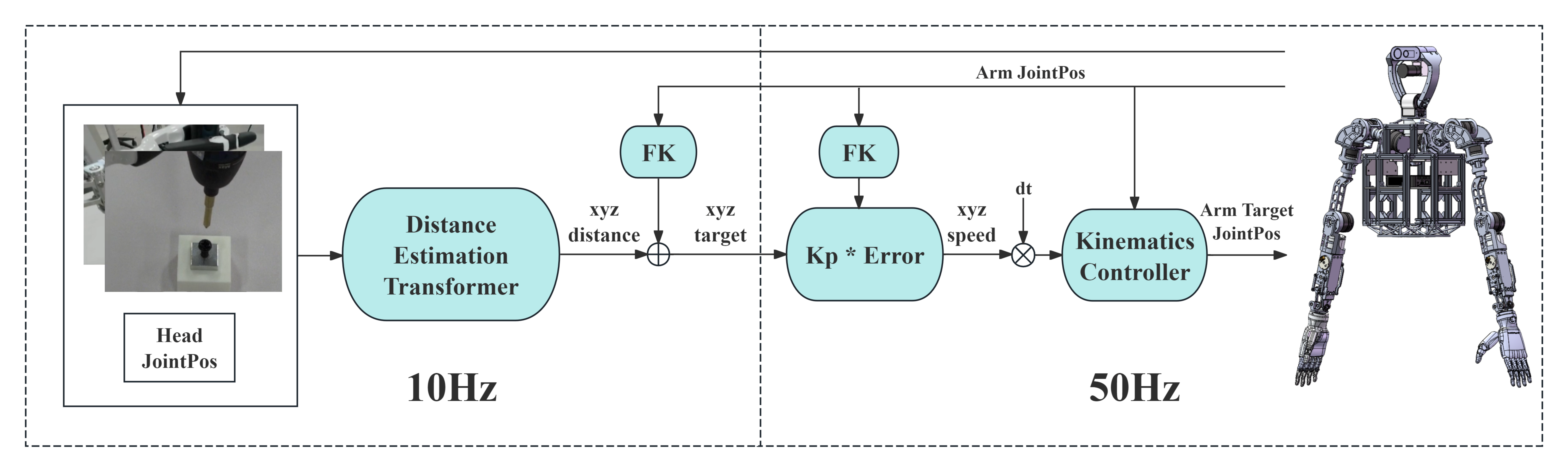}
    \caption{Framework of the control method.}
    \label{fig:controller}
\end{figure*}

To accomplish high-precision operation tasks, a desired IBVS system needs to acquire real-time images from the cameras and angles from the joint motors, then compute the relative position between the handheld tool and the target object, and eventually drive this discrepancy to zero through feedback control. 
Due to the limited computation resource onboard, a dual-frequency control architecture is designed as illustrated in Fig.~\ref{fig:controller}. The system first employs a Distance Estimation Transformer (DET) network at $10$ Hz to estimate the translational distance between tool and object, which can result in the end-effector's target position during each estimation. Subsequently, the difference between current and target positions is scaled by a proportional coefficient to yield the desired motion velocity. Finally, a nonlinear optimization method is employed to achieve temporally smooth joint motion based on inverse kinematics, driving the end-effector to its updated target position at $50$ Hz. The equivalence of the end-effector's motion and the tool's motion will be demonstrated in subsection~\ref{Data Collection}. 

\subsection{Distance Estimation Transformer}

\begin{figure*}[!t]
    \centering
    \includegraphics[width=1.66\columnwidth]{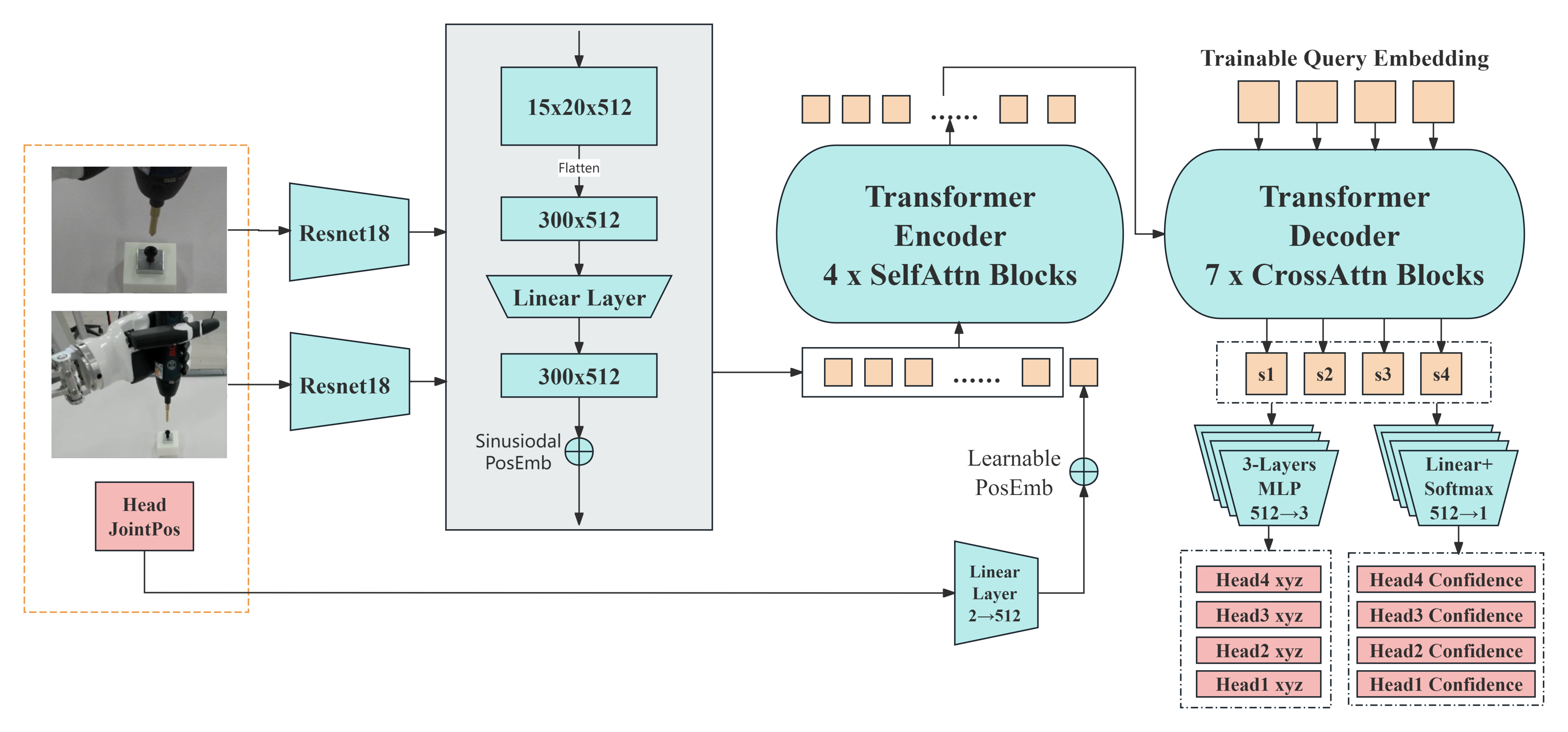}
    \caption{Architecture of Distance Estimation Transformer. }
    \label{fig:det}
\end{figure*}

As illustrated in Fig.~\ref{fig:det}, Distance Estimation Transformer is a neural network architecture designed for estimating the relative position between handheld tools and target objects. It takes real-time images from the head and torso cameras along with the head joint angles as input.
Inside DET, each image is first processed by a ResNet18~\cite{he2016deep} to obtain a feature map, which is then flattened to get a sequence of features. These features are subsequently projected to the embedding dimension with a linear layer. A 2D sinusoidal position embedding is added to preserve the spatial information. Guided by IBVS principles, system robustness against calibration errors and mechanical structure inaccuracies is enhanced by ignoring arm joints. Thus only head joints that can affect camera orientation are considered, namely head yaw and pitch movements. The angle data from these two joints undergo projection into an embedding space via a linear neural network layer, added with a learnable positional embedding. Finally, the combined representation serves as input to the Transformer encoder and is encoded into memory through multi-layer self-attention mechanism.

In RGB imagery, when position discrepancy exists between a screwdriver tip and a screw head, even minor misalignment can be visually discernible and yield directional motion determination. 
During multi-distance training, farther samples disproportionately influence network convergence compared to closer ones. 
At a farther distance, RGB data provide limited depth cues, leading to substantial estimation errors and consequently higher loss magnitudes, while at a closer distance, estimation errors typically induce relatively small loss values. 
This creates suppression effect that degrades close-range precision - a critical issue in robotic applications where high precision after convergence is desired, like the tiny object alignment task discussed in this paper.
To address this issue, distance-dependent output scaling is implemented to counteract the dominance of farther samples. This approach enables the network to extract refined features of positional discrepancy that are critical for precise control.

DET also features a decoder architecture employing several distinct queries to generate corresponding perception heads through Transformer decoding (Fig.~\ref{fig:det}). Each perception head maintains full attention to encoded image and joint angle token sequences. The output features of the decoder are processed by dedicated Multi-Layer Perceptrons (MLPs) to output amplified relative Cartesian positions with unique amplification coefficient optimized for each specific distance range. A linear neural network layer with softmax activation is also deployed for each output feature to determine whether the magnitude of distance is within the operational range. This mechanism is proposed as Multi-Perception-Head (MPH) in this paper. 


\subsection{Details of Multi-Perception-Head}
\label{Multi-Perception-Head Technique}

The output gains of the perception heads in MPH should correlate to confidence intervals. Each confidence interval is defined with mean value $\mu$ and range $\left [ \mu-\sigma, \mu+\sigma \right )$. A perception head is defined to achieve maximum confidence when the norm of a distance vector aligns with the confidence interval's mean value. With neural networks that are trained with batch normalization and L2 regularization, normalizing amplified output vectors with respect to unit vectors proves theoretically consistent. Guided by this framework, perception head output gain is set as $1/\mu$. Despite the effect of loss weighting mechanisms suppressing long-distance over short-distance heads, outputs are still empirically clamped to $\left [-3, 3 \right ]$ to prevent pathological learning from large output residual values.
Since the confidence intervals of each perception head are complementary, the confidence output is designed as a multiclass classification problem.
Define $\mathbf{d}_r$ as the real distance from the ground-truth dataset, then the ground-truth distance output $\mathbf{d}_o$ can be expressed as
\begin{align}
    \label{d_o}
    \mathbf{d}_o=clip(\mathbf{d}_{r} / \mu, -3, 3)
\end{align}
and the confidence output can be expressed as
\begin{align}
\label{con}
    con_o=\begin{array}{l} 
  \left\{\begin{matrix} 
  0,\left | \mathbf{d}_{r} \right | \notin \left [ \mu-\sigma, \mu+\sigma \right ) \\ 
  1,\left | \mathbf{d}_{r} \right | \in \left [ \mu-\sigma, \mu+\sigma \right )
\end{matrix}\right.    
\end{array}
\end{align}

Ideally, if the neural network precisely estimates the distance, the loss function can theoretically be designed such that the distance loss function is L1 loss within the confidence interval and 0 outside the confidence interval. However, in practice, although the confidence intervals for different outputs do not overlap, the network is very likely to make incorrect judgments near the boundaries of the confidence interval. Therefore, the weight of the distance loss function should smoothly decrease to $0$ at the boundaries of the confidence interval, rather than dropping abruptly to $0$.
To achieve this, the loss function weight is redesigned as follows, referred to as Gaussian Clipped Weight (GCW),
\begin{align}
    \label{gcw}
    gcw_h(x)=min(e^{\frac{1}{2}\alpha_h^2}e^{-\frac{1}{2}(\alpha_h\frac{x-\mu_h}{\sigma_h})^2},1)
\end{align}
where $\mu_h$ and $\sigma_h$ represent the mean and standard deviation in the Gaussian function. In the context of this paper, they describe the confidence interval of perception head $h$, defined as $\left [ \mu_h-\sigma_h , \mu_h+\sigma_h \right ]$. $\alpha_h$ is a hyperparameter that affects the concentration of the perception head. A larger value of $\alpha_h$ results in faster weight decay outside the confidence interval, making the perception head $h$ less sensitive to observations far away from the confidence interval. According to~\eqref{gcw}, the weight is $1$ within the confidence interval, and starts to decay from $1$ outside the confidence interval, but it will never drop below $0$. In addition, the coefficients of the Gaussian function are adjusted to make GCW a continuous function, such that sudden changes in weights at the boundaries of the confidence interval can be avoided.

Furthermore, set the number of perception heads as $n_h$, the training batch size as $n$, and $k$ as a hyperparameter describing the weights between distance loss and confidence loss, the loss function could then be defined as  
\begin{align}
\label{loss}
  \begin{aligned}
    \mathcal{L}&=\frac{1}{n}\sum_{i=1}^{n}(\mathcal{L}_{CE}(con_o^i, \hat{con_o^i})\\&+k\sum_h^{n_h}gcw_h(\left | \mathbf{d}_r\right |)\cdot \mathcal{L}_{L1}(\mathbf{d}_o^i, \hat{\mathbf{d}_o^i}))
\end{aligned}
\end{align}
where $\mathcal{L}_{L1}$ is the L1 loss function, $\mathcal{L}_{CE}$ is the multiclass cross entropy loss function and the $\hat{}$ sign indicates the variable is the output estimates from network. 
Note that during deployment, after network inference, the perception head with the highest confidence is selected and the head's distance output is multiplied by the corresponding $\mu$ to obtain the estimated distance.

\subsection{Data Collection}
\label{Data Collection}

For each set of input images, obtaining corresponding ground truth data is critical. While manual measurement remains theoretically viable, the extensive workload required for large-scale data collection renders this approach impractical. To address this challenge, the robot's kinematics is employed for automated ground truth data acquisition. The torso coordinate frame is selected as the reference frame,
given that the torso camera maintains a fixed pose with respect to the torso, while the pose of the head camera is controlled by the yaw and pitch joints of the head - both serving as network input. As a fundamental prerequisite to justify the data collection process, the mathematical equivalence between tool and end-effector displacement within the torso coordinate system is first demonstrated.

Consider a robot manipulating a tool moving from position $a$ to position $b$, where the tool's pose relative to the torso frame transforms from $_{ta}\mathbf{T}$ to $_{tb}\mathbf{T}$, and the end-effector's pose changes from $_{ea}\mathbf{T}$ to $_{eb}\mathbf{T}$. Within the torso coordinate system, let ${\mathbf{T}}_{dt}$ represent the pose transformation between $_{ta}\mathbf{T}$ and $_{tb}\mathbf{T}$, and ${\mathbf{T}}_{de}$ denotes the pose transformation between $_{ea}\mathbf{T}$ and $_{eb}\mathbf{T}$, which is 
\begin{align}
\label{b_T}
\begin{cases}
    _{tb}{\mathbf{T}} = {\mathbf{T}}_{dt} \cdot {_{ta}\mathbf{T}} \\
    _{eb}{\mathbf{T}} = {\mathbf{T}}_{de} \cdot {_{ea}\mathbf{T}}
\end{cases}
\end{align}
On the other hand, during data collection, consistent tool-grasping configuration is maintained within each measurement group. This requires identical tool pose relative to the end-effector coordinate frame $^e_t{\mathbf{T}}$ at both positions $a$ and $b$, which can be mathematically written as
\begin{align}
\label{t_T}
\begin{cases}
    {_{ta}\mathbf{T}}={_{ea}\mathbf{T}}\cdot {^e_t{\mathbf{T}}}\\
    {_{tb}\mathbf{T}}={_{eb}\mathbf{T}}\cdot {^e_t{\mathbf{T}}}
\end{cases}
\end{align}
Combining (\ref{b_T}) and (\ref{t_T}), it can be seen that
\begin{align}
\label{T_dt=T_de}
    {\mathbf{T}}_{dt} = {\mathbf{T}}_{de}
\end{align}

Considering that the homogeneous transformation matrix $\mathbf{T}$ can be decomposed into the rotation matrix $\mathbf{R}$ and the translation vector $\mathbf{p}$, combining (\ref{b_T}) and (\ref{T_dt=T_de}) can yield the translation of the tool from $a$ to $b$ as
\begin{align}
\label{p_de with R}
    _{tb}{\mathbf{p}}-{_{ta}{\mathbf{p}}}=\mathbf{R}_{de} \cdot {_{ta}{\mathbf{p}}}+\mathbf{p}_{de}-{_{ta}{\mathbf{p}}}
\end{align}
It can be seen that within the torso coordinate system, the computed translational distance between tool and object is influenced by their relative rotational error. However, in the proposed control framework, the absolute position of the tool relative to the torso frame origin $ {_{ta}{\mathbf{p}}}$ remains an unknown quantity. Consequently, precise determination of the position between tool and object would be unattainable in the presence of rotational misalignment.
Therefore, as the initial attempt to tackle this problem, this study assumes that the tool and object have been pre-aligned rotationally within task-acceptable tolerance. 
During data acquisition, consistent end-effector rotational orientation within each measurement group is ensured. Under these conditions, the rotation matrix $\mathbf{R}_{de}$ in (\ref{p_de with R}) reduces to identity matrix $\mathbf{I}$, which yields
\begin{align}
\label{p_de}
    {_{tb}{\mathbf{p}}}-{_{ta}{\mathbf{p}}}=\mathbf{p}_{de}={_{eb}{\mathbf{p}}}-{_{ea}{\mathbf{p}}}
\end{align}
Therefore, the equivalence of the movement of handheld tool to the end-effector's trajectory in the designated reference frame is confirmed, thereby enabling reliable automated ground truth generation.

The data collection process proceeds as follows: Initially, the robot with the tool is teleoperated to achieve precise alignment with the target object, ensuring both the screwdriver tip and the screw head remain within the field of view of both cameras. The corresponding images, joint angles and end-effector pose (computed via forward kinematics) are recorded as benchmark data. Subsequently, while maintaining fixed end-effector rotation and object grasping configuration, the robotic arm is randomly translated within the operational space and the head joints are also randomly rotated, to generate images, joint configurations and end-effector poses as data points. Each measurement group consists of one benchmark paired with multiple associated data points. Following (\ref{p_de}), the actual tool-to-object distances within each group are computed by subtracting the benchmark end-effector position from the corresponding positions at different data points. Among different groups, controlled variation is introduced by randomly repositioning the target object, altering tool grasping configuration, and applying bounded random rotational errors between tool and object to simulate imperfect pre-alignment conditions. Eventually, the data are processed using (\ref{d_o}), (\ref{con}) and (\ref{p_de}) to generate the final training dataset. 

Note that although the data collection process utilizes robot kinematics, the relative position between the tool and target object is obtained through the difference between two similar kinematic calculations. Particularly, this work focuses on position precision at close range, where calibration errors from the robot's zero-point calibration are largely cancelled out through the subtraction, with the remaining uncompensated portion being a second-order small quantity that can be reasonably neglected. Therefore, this data collection procedure is robust against zero-point calibration errors and aligns with the IBVS philosophy.

\subsection{Kinematics Controller}
\label{IK}

Since the movement of the handheld tool and the end effector have shown to be equivalent in this paper, the end-effector's target position can be updated using DET at 10Hz. Then, the error between the end-effector's current and target positions is calculated to get the desired end-effector velocity at 50 Hz. The velocity is further integrated to yield an intermediate target position command for the end-effector. This process ensures a smooth and rotationally invariant desired motion trajectory at every iteration. To approach the target position, 
a nonlinear optimization method is employed to complete the inverse kinematics calculation and generate joint position commands for the robotic arm. The optimization utilizes current joint angles as the initial seed for solving the problem formulated as
\begin{align}
\label{nlp for ik}
\begin{array}{cc}
\min _{q} & \left[ fk(\mathbf{q})-\hat{\mathbf{y}} \right]^2 \\
\text { s.t. } &\mathbf{lb} \leq \mathbf{q} \leq \mathbf{ub},
\end{array}
\end{align}
where $fk$ is the forward kinematics of the robot, $\hat{\mathbf{y}}$ is the target pose, and $\mathbf{lb}$ and $\mathbf{ub}$ describe the lower and upper bounds of joint angles. The optimization is solved using the sequential least squares programming (SLSQP) algorithm implemented using the library \textbf{NLopt}~\cite{NLopt}, while the forward kinematics and Jacobian matrix are solved by the library  \textbf{Pinocchio}~\cite{carpentier2019pinocchio}. 



\section{EXPERIMENTS}
\label{RESULT}

\subsection{Experimental Setup}
\label{Experiment Setup}

\begin{figure}[!t]
    \centering
    \subfigure[\label{fig:task}]{
        \includegraphics[height=.35
        \columnwidth]{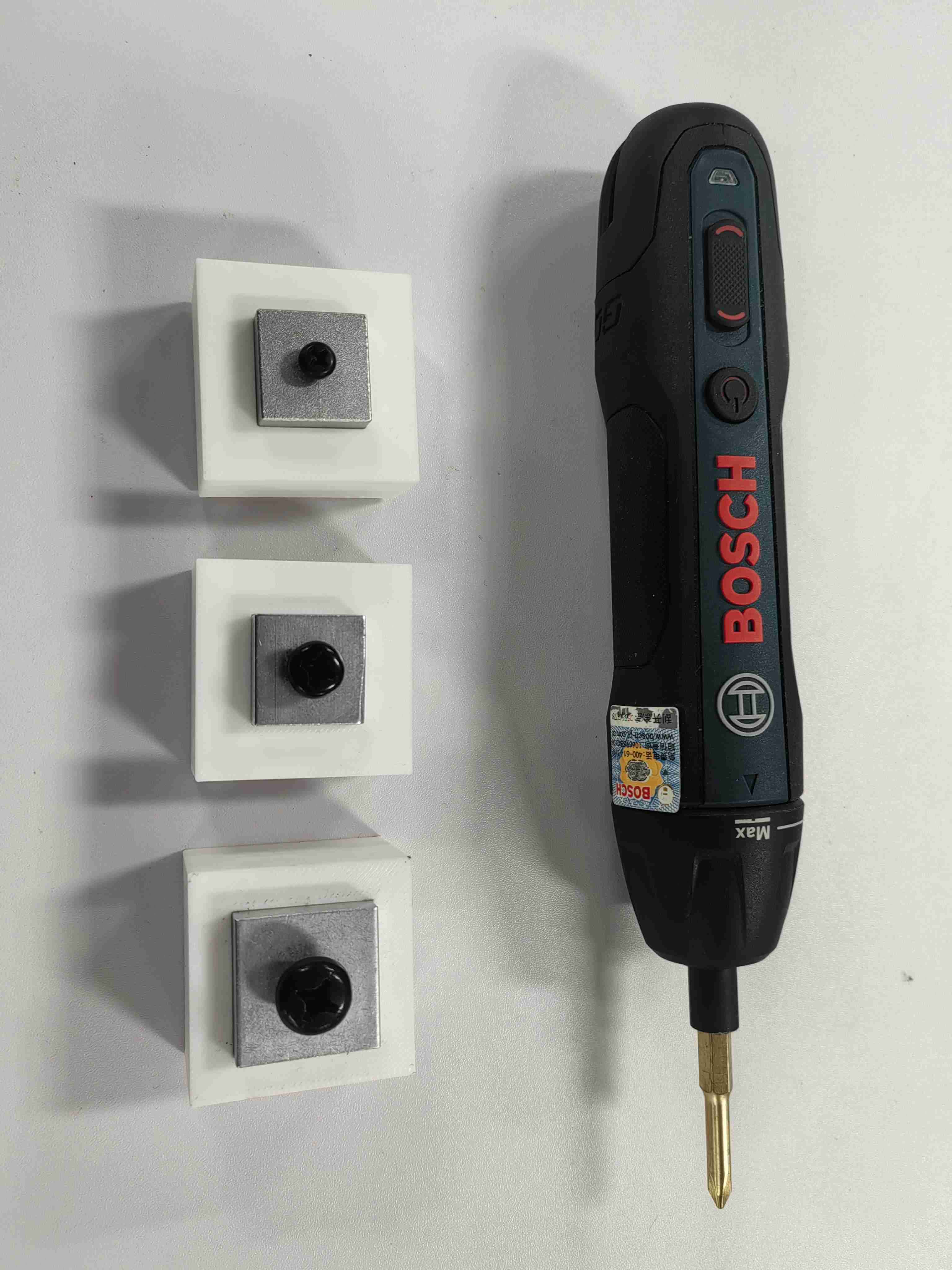}
    }
    \hfill
    \subfigure[\label{fig:taskstart}]{
        \includegraphics[height=.35
        \columnwidth]{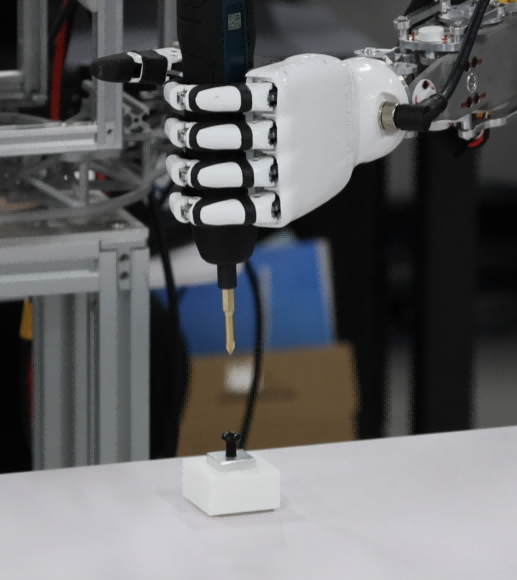}
    }
    \hfill
    \subfigure[\label{fig:taskend}]{
        \includegraphics[height=.35
        \columnwidth]{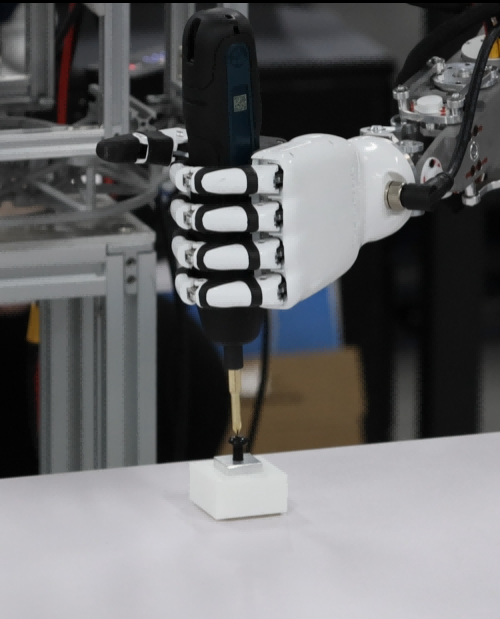}
    }
    \caption{Experiments overview: (a) The electric screwdriver and the test fixtures equipped with M8, M6, and M4 screws for the experiments. (b) The initial state of a single experimental trial, where a significant spatial separation is shown between the screwdriver tip and the screw head center. (c) The final state of the trial, where the screw head center and the screwdriver tip are aligned. }
    \label{fig:task_description}
\end{figure}

As mentioned, the purpose of the experiments is to evaluate the proposed visual servoing method on enabling a robotic arm to precisely align a handheld electric screwdriver tip with the screw head slot of various sizes (M4, M6, and M8 specifications), as illustrated in Fig.~\ref{fig:task_description}. To quantify the performance, two metrics are employed: success rate (SR) in 100 trials and convergence error (CE). A trial is considered successful if after the controller achieves convergence, the screwdriver tip can physically engage with the screw head slot upon applying downward pressure. The convergence error is quantified by the manually measured Euclidean distance between the screwdriver tip and the slot center. For comparative analysis, we also implement another two baseline models: a DET network with a Single-Perception-Head (SPH) mechanism and a conventional ResNet18 architecture with an MLP head.

For each experimental task, 40 measurement groups are collected, with each group comprised of one ground truth and 15-25 associated data points. Data from different tasks are used together for training to improve robustness and generalization. Given the processing capability of neural networks and the required Field of View (FOV) during operation,
the raw images from both cameras are preprocessed by first resizing the raw images to 1280×960 pixels and then extracting the central 640×480 pixels as the input to the neural network. Thus, the effective FOV of the two cameras is approximately half of the initial FOV, which balances the competing high spatial resolution requirements for precise feature localization and the sufficient contextual information for robust visual servoing. 
Next, the collected measurement groups undergo preprocessing based on \ref{Data Collection} to generate the training dataset. During training, the AdamW optimizer is employed with a fixed learning rate of $10^{-5}$ over 125 epochs. Each epoch represents a complete iteration through the entire dataset, ensuring thorough exposure to all collected samples.

\begin{figure}[t]
    \centering
    \includegraphics[width=0.9\columnwidth]{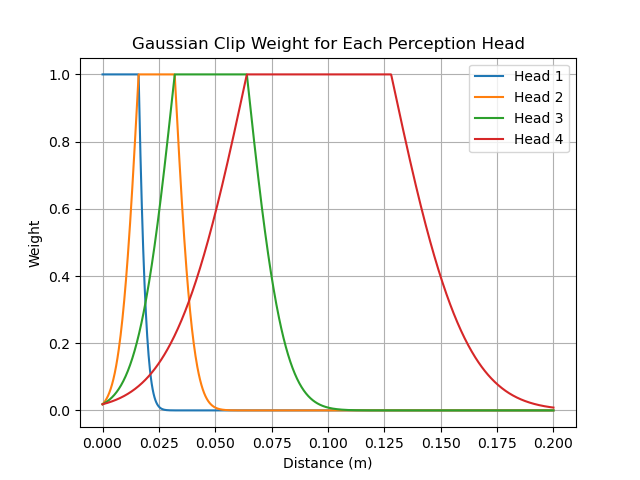}
    \caption{The variation of Gaussian Clip Weight for each perception head. The weight is 1 within the confidence interval, and starts to decay outside the confidence interval. A zero distance corresponds to a $3\sigma$ deviation in Head2-Head4. }
    \label{fig:gcw}
\end{figure}

For DET, the number of perception heads is set to 4. The hyperparameters of each perception head can be found in Table~\ref{head hyperparameters}. The variation of the loss weight defined by the GCW function based on distance is shown in Fig.~\ref{fig:gcw}. The confidence interval of Head2-Head4 is designed as a geometric sequence, while the confidence interval of Head1 is the minimum interval complementary to Head2. A zero distance corresponds to a $3\sigma$ deviation in Head2-Head4. The decay coefficient $\alpha$ of Head1 is set to 1.6, larger than those of Head2-Head4, to ensure its output peak to have a relatively small weight. This design makes each perception head still sensitive to observations slightly beyond the confidence interval of length $2\sigma$, ensuring a balance between the perception head's concentration and robustness to misselection.

\begin{table}[h]
\centering
\caption{ The hyperparameters of each perception head. }
\label{head hyperparameters}
\begin{tabular}{lcccccc}
\toprule
Name & Confidence Interval (m) & $\mu$ (m) & $\sigma$ (m) & $\alpha$ \\
\midrule
Head1     & $[0, 0.016)$     & 0.008    & 0.008       & 1.6      \\
Head2     & $[0.016, 0.032)$ & 0.024    & 0.008       & 1        \\
Head3     & $[0.032, 0.064)$ & 0.048    & 0.016       & 1        \\
Head4     & $[0.064, 0.128)$ & 0.096    & 0.032       & 1        \\
\bottomrule
\end{tabular}
\end{table}

\subsection{Results and Discussion}



The results from experiments are shown in Table~\ref{tab:result}. It can be seen that the proposed method demonstrates promising success rates and precision. As the screw size decreases from M8 to M4, the same visual distinctness of the positional discrepancy between the screwdriver tip and the screw head center becomes progressively more pronounced in image space, resulting in a reduction in mean alignment error. However, the smaller dimension of the M4 screw imposes stricter error tolerance, leading to a reduced success rate in the alignment task. 
The experimental results on the two baseline models can also be found in Table~\ref{tab:result}. They reveal that the proposed method achieves significantly better performance, with 16\% higher average success rate and 25\% better precision compared to the DET-SPH baseline, and 47\% higher average success rate and 54\% better precision compared to the ResNet18+MLP baseline. The results substantiate the effectiveness of both the DET architecture and the MPH mechanism in enhancing visual feature extraction and control precision for tiny object alignment tasks.

\begin{figure}[!t]
    \centering
    \subfigure[
        \label{fig:train_loss}
    ]{
        \includegraphics[height=.35
        \columnwidth]{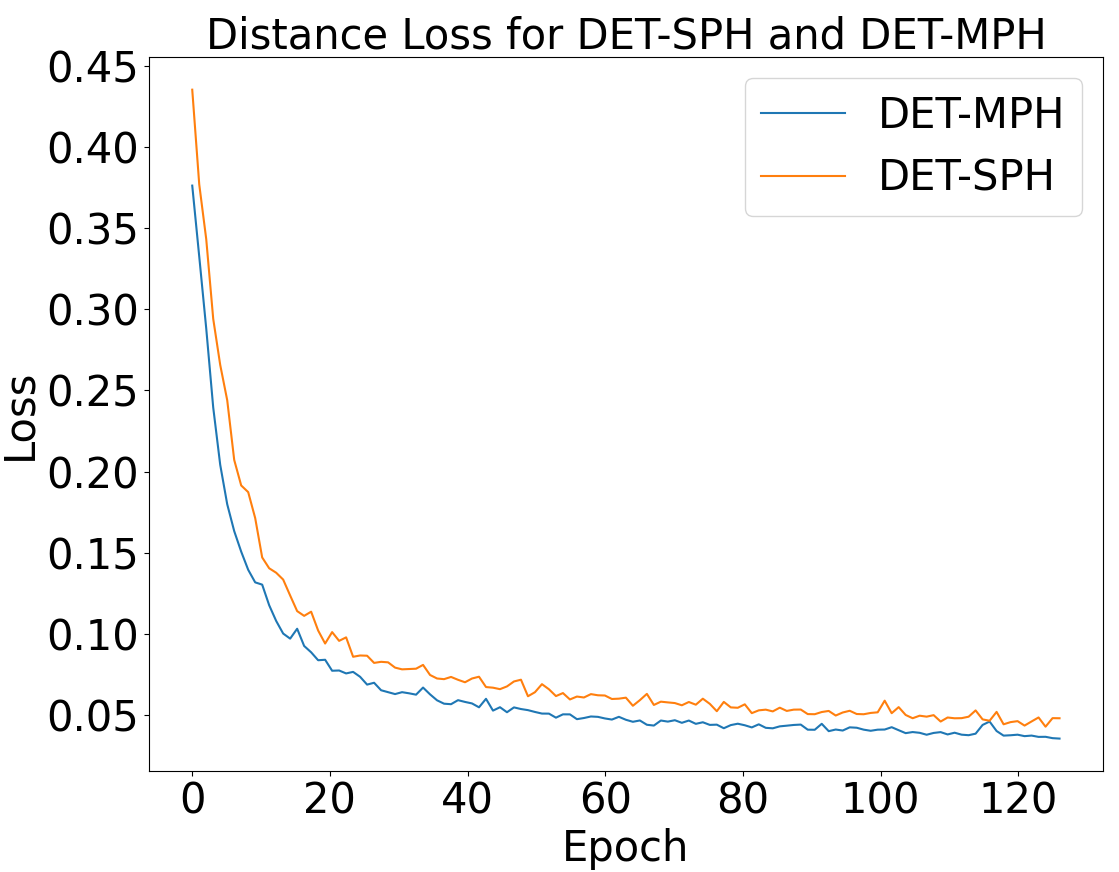}
    }
    \hfill
    \subfigure[
        \label{fig:train_err}
    ]{
        \includegraphics[height=.35
        \columnwidth]{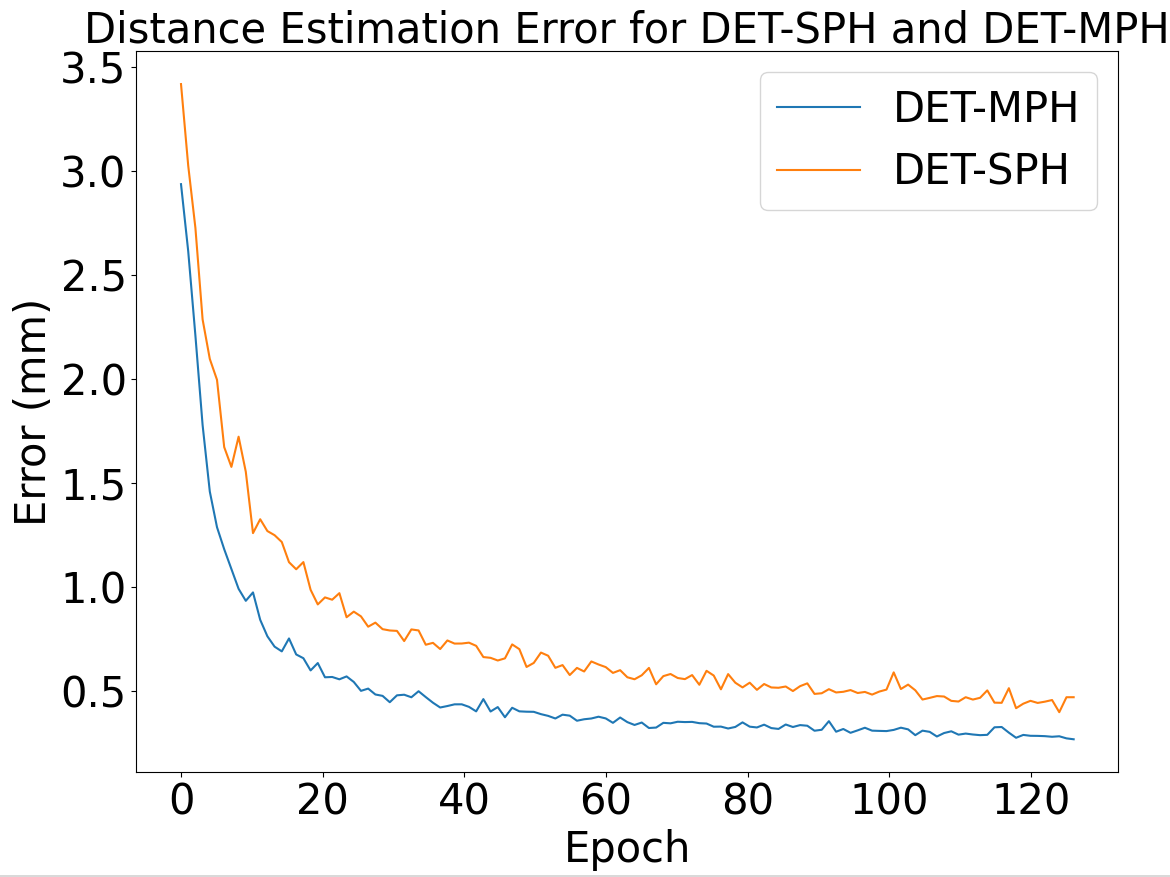}
    }
    \caption{The convergence process of distance losses and estimation errors at close distance for DET-SPH and DET-MPH during training. (a) The convergence process of distance loss for DET-SPH and DET-MPH, where a fixed output gain of $20m^{-1}$ is set for DET-SPH, and the output gain and loss weights of DET-MPH are described in Subsection \ref{Experiment Setup}. (b) The average estimation errors of DET-SPH and DET-MPH for samples with distances less than $0.016m$.  All of these samples fall within the confidence interval of Head1 in DET-MPH.}
    \label{fig:Loss_and_Err}
\end{figure}

\begin{table}[ht]
\vspace{2mm}
\caption{Performance of DET-MPH, DET-SPH and ResNet18+MLP on M8/M6/M4 Tasks: \\Success Rate (SR) and Convergence Error (CE) over 100 Trials. }
\label{tab:result}
\begin{center}
\setlength{\tabcolsep}{0.65mm}{
\begin{tabular}{lcccccccc}
\toprule
\multirow{2}{*}{} & \multicolumn{2}{c}{M8} & & \multicolumn{2}{c}{M6} & & \multicolumn{2}{c}{M4} \\ \cmidrule{2-3} \cmidrule{5-6} \cmidrule{8-9} 
                  & SR       & CE (mm)     & & SR       & CE (mm)     & & SR       & CE (mm)     \\
\midrule
\textbf{DET-MPH(ours)}        & \textbf{100\%}    & \textbf{1.3 $\pm$ 0.6}    & & \textbf{100\%}    & \textbf{1.0 $\pm$ 0.4}    & & \textbf{93\%}     & \textbf{0.8 $\pm$ 0.5}    \\
DET-SPH         & 85\%         & 1.5 $\pm$ 0.6            & &  89\%        & 1.3 $\pm$ 0.6            & &  71\%        & 1.3 $\pm$ 0.5            \\
Resnet18+MLP      & 57\%         & 2.4 $\pm$ 1.3            & &  50\%        & 2.4 $\pm$ 0.6            & & 44\%         &  2.0 $\pm$ 1.0          \\
\bottomrule
\end{tabular}}
\end{center}
\end{table}

Fig.~\ref{fig:Loss_and_Err} illustrates the convergence of distance loss and estimation error for close-range samples during training. Fig.~\ref{fig:train_loss} shows that the distance loss of DET-MPH converges slightly faster than DET-SPH, but there is no significant difference overall. However, as shown in Fig.~\ref{fig:train_err}, the estimation error of DET-MPH converges significantly faster than that of DET-SPH for samples with distances less than $0.016m$. Specifically, DET-MPH achieves an estimation error of below $0.001m$ within only 8 epochs during training for the close-range samples, whereas DET-SPH requires nearly 20 epochs to reach the same level of performance. This indicates that the larger loss contribution from distant samples in DET-SPH suppresses its ability to process close-range samples effectively. In contrast, DET-MPH achieves a more balanced training across the whole operational range, enabling faster convergence at close distances. Furthermore, the higher gain in DET-MPH at close range results in a smaller average error. As depicted in Fig.~\ref{fig:train_err}, the estimation error of DET-SPH at close distance is approximately $1.6$ times that of DET-MPH upon training convergence.

\begin{figure}[!t]
    \centering
    \subfigure[
        \label{fig:failexp}
    ]{
        \includegraphics[width=.46
        \columnwidth]{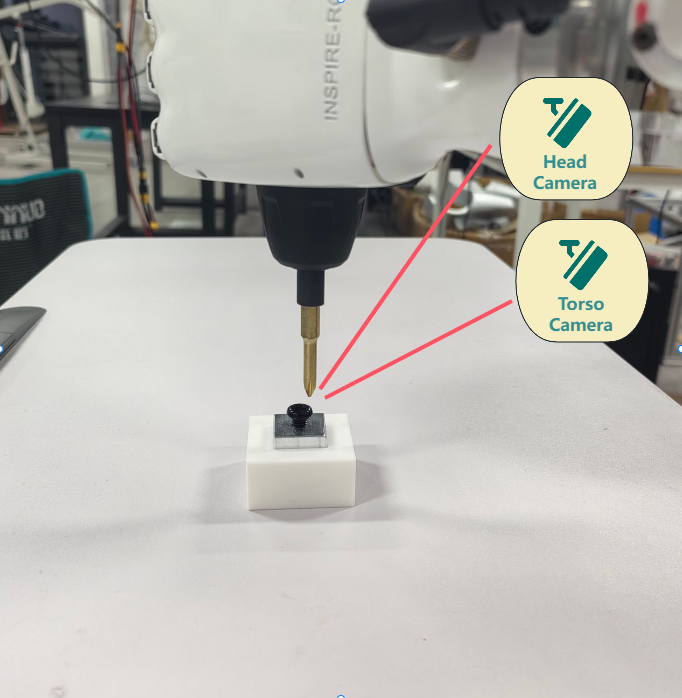}
    }
    \hfill
    \subfigure[
        \label{fig:successexp}
    ]{
        \includegraphics[width=.45
        \columnwidth]{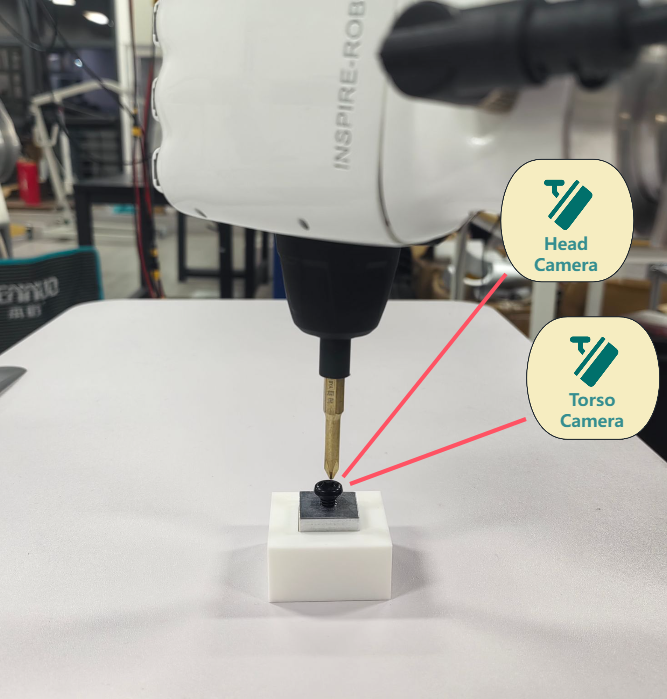}
    }
    \caption{A typical case where DET-MPH performs better than DET-SPH network. (a) A typical failure case in DET-SPH network experiments. The screwdriver tip is on the line connecting the head camera and the screw slot center. (b) DET-MPH guides the screwdriver tip to achieve superior alignment with the slot center.}
    \label{fig:typicalcase}
\end{figure}

Fig.~\ref{fig:failexp} illustrates a typical failure case observed in experiments with DET-SPH , where the screwdriver tip is on the line connecting the head camera and the screw head center. The head and torso cameras both perceive that the screwdriver tip has approximately centered over the screw head, causing the network to falsely conclude proper alignment. On the contrary, the DET-MPH architecture's enhanced gain at close range amplifies such proximity error. Substantial training loss is generated to drive the network to discern subtle positional discrepancy in image space, which ultimately guides the screwdriver tip to achieve superior alignment with the slot center, as demonstrated in Fig.~\ref{fig:successexp}. These results demonstrate that the MPH mechanism significantly improves precision even under the hardware's limitations, achieving enhanced performance without additional cameras or modified visual configurations.

\section{CONCLUSIONS}
\label{CONCLUSION AND LIMITATION}
This paper presents a novel vision-based framework for high-precision tiny object alignment using humanoid robots. The core contribution is the DET architecture that effectively fuses multiple camera images with joint angles to achieve precise visual servo control, without relying on manually designed visual features or markers. The MPH design with gain scheduling successfully overcome the challenge of achieving high precision across a wide operational range. 
Besides, a kinematics-based data collection methodology is used to reduce the labor for gathering training data. The proposed method demonstrates an average positional precision of $0.8$-$1.3$ mm and a success rate of $93$\%-$100$\% for the alignment tasks with M4-M8 screws, showing a significantly better performance than frameworks with either single perception head or traditional convolutional neural network.

However, the method is fundamentally constrained by the inherent coupling between translational and rotational transformations, necessitating the assumption of pre-existing rotational alignment between tool and target. This assumption, though operationally necessary in our current implementation, may be difficult to achieve in numerous real-world scenarios, potentially limiting the method's practical applicability. In fact, different from translational error which shows more obvious features in RGB images, rotational error is often hard to identify from certain directions. In contrast, 3D point cloud may be more suitable for measuring rotational errors. Future work will focus on developing a more generalized alignment strategy and ultimately achieve high-precision alignment in both translational and rotational domains.







\bibliographystyle{IEEEtran}
\bibliography{ref}

\end{document}